\documentclass[conference]{IEEEtran}
\IEEEoverridecommandlockouts

\usepackage{amsmath,amssymb,amsfonts}
\usepackage{adjustbox}
\usepackage[ruled,vlined]{algorithm2e}
\usepackage{amssymb}
\usepackage{arydshln}
\usepackage{array}
\usepackage{balance}
\usepackage{booktabs}
\usepackage{cite}
\usepackage{colortbl}
\usepackage{comment}
\usepackage{enumitem}
\usepackage{etoolbox}
\usepackage{graphicx}
\usepackage{hyperref}
\usepackage{listings}
\usepackage{multicol}
\usepackage{multirow}
\usepackage{nameref}
\usepackage{soul}
\usepackage{capt-of}
\usepackage{tabularx}
\usepackage{textcomp}
\usepackage{threeparttable} 
\usepackage{tikz}
\usepackage{url}
\usepackage{xcolor}

\def\BibTeX{{\rm B\kern-.05em{\sc i\kern-.025em b}\kern-.08em
    T\kern-.1667em\lower.7ex\hbox{E}\kern-.125emX}}

\def\eg{\emph{e.g., }} 

\def\ie{\emph{i.e., }}

\newcommand{\insertfig}{\vspace{1em}\includegraphics[width=\linewidth]{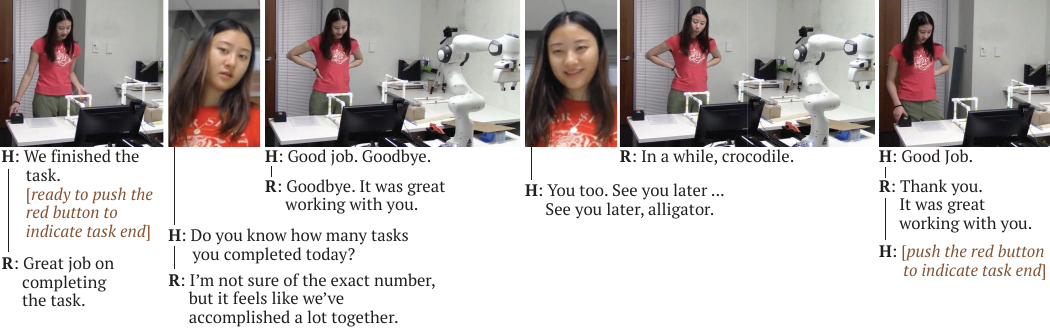}\captionof{figure}{We investigate users’ interactions with an autonomous manipulator robot that engages in small-talk during a collaborative task. 
The example here shows the participant engaged in a \textit{lingering conversation} with the robot after the task completion. }
\label{fig:teaser}}

\makeatletter
\apptocmd{\@maketitle}{\centering\setcounter{figure}{0}\insertfig}{}{}
\makeatother

\begin{document}

\title{``\textit{See You Later, Alligator}'': 
Impacts of Robot Small Talk on 
Task, Rapport, and Interaction Dynamics
in Human-Robot Collaboration


\thanks{
\textbf{\textit{CRediT Author Statement---}}
\textbf{Pineda}: Conceptualization, Methodology, Software, Writing, Formal Analysis, Data Curation, Visualization, Project Administration. 
\textbf{Brown}: Validation, Investigation, Formal Analysis, Writing - Original draft.
\textbf{Huang}: Conceptualization, Methodology, Writing - Review \& Editing, Visualization, Supervision, Funding acquisition. 
}
\thanks{
\textbf{\textit{AI Use---}}
Text edited with LLM; output checked for correctness by authors.
}

}

\author{\IEEEauthorblockN{Kaitlynn Taylor Pineda}
\IEEEauthorblockA{
\textit{Johns Hopkins University}\\
Baltimore, MD, USA \\
kpineda3@jhu.edu}

\and

\IEEEauthorblockN{Ethan Brown}
\IEEEauthorblockA{
\textit{Johns Hopkins University}\\
Baltimore, MD, USA \\
ebrow170@jhu.edu}

\and

\IEEEauthorblockN{Chien-Ming Huang}
\IEEEauthorblockA{
\textit{Johns Hopkins University}\\
Baltimore, MD, USA \\
cmhuang@cs.jhu.edu}
}

\maketitle

\begin{abstract} 
Small talk can foster rapport building in human-human teamwork; 
yet how non-anthropomorphic robots, such as collaborative manipulators commonly used in industry, may capitalize on these social communications remains unclear. 
This work investigates how robot-initiated small talk influences task performance, rapport, and interaction dynamics in human-robot collaboration. 
We developed an autonomous robot system that assists a human in an assembly task while initiating and engaging in small talk.
A user study ($N = 58$) was conducted in which participants worked with either a \textit{functional} robot, which engaged in only task-oriented speech, or a \textit{social} robot, which also initiated small talk. 
Our study found that participants in the social condition reported significantly higher levels of rapport with the robot.
Moreover, all participants in the social condition responded to the robot's small talk attempts; 59\% initiated questions to the robot, and 73\% engaged in lingering conversations after requesting the final task item. 
Although active working times were similar across conditions, participants in the social condition recorded longer task durations than those in the functional condition.
We discuss the design and implications of robot small talk in shaping human-robot collaboration.


\end{abstract}


\section{Introduction}
Human collaboration is more than task execution and typically includes off-task social interactions that facilitate rapport building, a vital element for sustained collaboration \cite{kopp2021success}. 
Indeed, small talk is a common social behavior used in human-human interactions to build rapport \cite{Laver1981LinguisticRA}, particularly in workplace settings.
While research has shown that robots with human-like features (\eg humanoid robots) can benefit from small talk to foster trust, rapport, and collaboration \cite{paradeda2016facial,seo2018investigating}, less is known about how non-anthropomorphic robots, such as industrial robotic arms, can capitalize on social interactions. 
Notably, these non-anthropomorphic robots are already widely deployed in industrial settings, performing highly functional tasks \cite{BenAri_Mondada_2018}; however, they are not designed for participating in off-task social interactions \cite{Welfare_Hallowell_Shah_Riek_2019} commonly seen in human co-workers.
As their role in workplaces expands, it is imperative to understand whether these robots can engage in small talk and foster rapport in human-robot teams.


Although prior research has shown that people enjoy social interactions in manufacturing settings and expect their robot co-workers to have social capabilities \cite{Sauppé_Mutlu_2015}, the robots studied have a human-like form factor, \ie the Baxter robot with animated facial expressions.
It remains unclear how people might engage with non-anthropomorphic manipulators, typically perceived as less social. 
This work investigates the impacts of a non-anthropomorphic robot engaging in social exchanges, specifically small talk, during a collaborative assembly task. 

We developed an autonomous robot system for task-oriented and social interactions, incorporating small talk in human-robot collaboration by utilizing a large language model to recognize a user's social intent and generate small talk responses.
Through a user study with 58 participants, we gathered empirical evidence showing that small talk fosters rapport, though it leads to longer task duration, highlighting the need to balance social interaction with task efficiency.
Notably, all participants who interacted with the social robot ($n = 29$) responded to its small talk attempts, with many ($n = 17$) continuing conversations even after completing all task-related requests.
We also found that participants working with the social robot exhibited more happy expressions during the interaction.
This work makes three key contributions: 
\begin{enumerate}[leftmargin=*]

\item We design and develop an autonomous robot system capable of engaging in naturalistic small talk with people.

\item We report empirical evidence showing effects of robot small talk on task efficiency, rapport building, and interaction experience.

\item We discuss insights on designing small talk for human-robot collaboration, as well as the ethical considerations of integrating LLMs into such interactions.
\end{enumerate}

\section{Background and Related Works}



Human-robot interaction research has explored roles that robots take on across various domains, 
highlighting the importance of social behaviors for fostering rapport, trust, and collaboration \cite{seok2022cultural, laban2020tell, seo2018investigating}. 
However, most robots in real-world applications today are non-anthropomorphic, such as collaborative robots (cobot) in the form of a manipulator. 
These robots are primarily designed for functional tasks in manufacturing environments, where they perform tedious, repetitive, or dangerous tasks while humans handle more complex ones \cite{BenAri_Mondada_2018}. 
Yet, how to design effective human-robot interaction for this type of collaborative robot to work alongside people is still an active research area.
While prior research has sought to understand how social features, such as gaze, breathing \cite{terziouglu2020designing}, 
playfulness, \cite{chowdhury2021you}, and dialogue \cite{pineda2024youmightlikeit}
may be introduced into 
collaborative robots, 
research on how social speech interactions might influence non-anthropomorphic collaborative robots is still limited.

\subsection{Small Talk in Humans}

Small talk, often referred to as light, casual conversation, plays a crucial role in building rapport, trust, and social bonds between individuals \cite{Laver1981LinguisticRA, coupland2014small}.
In workplace settings, small talk has been shown to improve relationships between co-workers and increase creativity by allowing new ideas to emerge during discussions \cite{Dooly_Tudini_2016}.
Small talk typically occurs at the start of interactions. 
It covers non-task-oriented topics such as weather and sport \cite{holmes2005small}, with people naturally signaling their conversation switch from small talk topics to work talk.
\cite{di_ferrante_transitioning_2021}.
Though small talk is commonly used in cultures like the U.S., perspectives on its importance can vary. 
For example, some Chinese professionals working in Australia may be unfamiliar with the role of small talk, while in Germany, greetings like “How are you?” are often interpreted literally, as casual small talk is not typically a cultural norm \cite{cui2015small, rings1994beyond, bickmore2005social}.

\subsection{Human-Machine Small Talk}

The integration of social speech in human-machine interactions has been explored with 
voice assistants \cite{mahmood2025user}, 
conversational agents \cite{clark2019good, feine2019taxonomy, kluwer2011-like, Ostrowski_Zygouras_Park_Breazeal_2021}, 
virtual agents \cite{zhou2019trusting}, 
and social robots \cite{Skantze_2017, nicholsHaruFriends2022,janssensBelpaemeCoolGlasses2022}.
These studies have shown that small talk can enhance user trust, engagement, and rapport, especially when combined with social cues like 
facial expressions \cite{paradeda2016facial}, 
emotional displays \cite{aroyo2018trust}, or gaze \cite{babel2021small}.
In a mock industrial scenario, researchers found that people naturally engage in rapport-building behaviors, such as friendly communication and cooperative gestures, when interacting with a collaborative robot \cite{seo2018investigating}.
However, this research and others \cite{laplazaIVORobot2022} primarily involve humanoid robots, which differ from typical industrial robots in appearance and function.
The application of small talk to foster rapport with non-anthropomorphic robots, not typically designed for conversation, has been less explored. 

Interviews with assembly-line workers revealed concerns that introducing robots into the workplace could reduce social interactions between colleagues, increase boredom, and fail to meet their desire for human conversation \cite{Welfare_Hallowell_Shah_Riek_2019}.
Furthermore, researchers have noted that robot operators have a desire for increased social interactions with the robots they work with, wishing they could engage in small talk with them similar to how they do with their co-workers \cite{Sauppé_Mutlu_2015}.
While our goal is not to replace human-human interactions with more social human-robot collaborations, 
we believe it is worthwhile to understand whether people will accept such machines and if these robots can still offer interaction benefits, like increased rapport, despite their non-anthropomorphic form.

\subsection{Leveraging LLMs for Robot Small Talk}

Prior work has developed a model blending task-oriented dialogue with small talk for robots in industrial settings, aiming to generate conversational responses to facilitate human-robot collaboration \cite{li2022tod}; however; this study focused more on evaluating user perceptions of the conversational models rather than on how users behave and respond to small talk from non-anthropomorphic robots. 
Another sequential work incorporated more dialogue into collaborative systems by creating a speech-enabled virtual assistant that leverages natural language processing and fine-tunes a BERT model to predict user intents across various industrial tasks \cite{li2023speech}.
However, this system focuses primarily on intent recognition without exploring the social dynamics that small talk can introduce. 

A more recent work has integrated a large language model (LLM) into a 7-DoF robotic arm with a face and multimodal sensory inputs, enabling both task-oriented and social interactions in human-robot collaboration \cite{allgeuer2024robots}.
The robot’s face facilitates social interactions, while the LLM is grounded in real-world data to execute physical tasks and social exchanges. 
While our system similarly utilizes an LLM to perform intent recognition and social response generation, in this work, we investigate how robot small talk affects interaction dynamics between a non-anthropomorphic manipulator and people.


\section{Engineering a Robot to Engage in Small Talk}
\label{sec:robot_system}
\begin{figure}
    \centering
    \includegraphics[width=\linewidth]{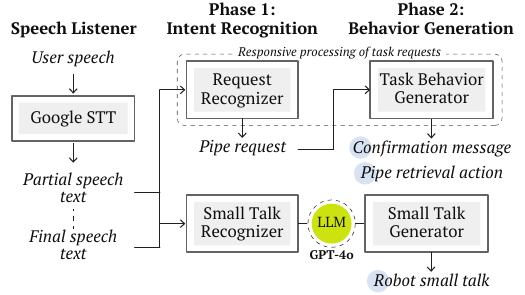}
    \caption{System Overview. The blue circles in Phase 2 denote robot behaviors.}
    \label{fig:system-diagram}
\end{figure}




We developed an autonomous robot system to assist people in a collaborative task while engaging in small talk. 
At a high level, our system includes 
a continuous \textit{speech listener},
an \textit{intent recognizer} that differentiates between task requests and social speech, 
and a \textit{behavior generator} that manages both physical and speech actions (Fig. \ref{fig:system-diagram}).
To support natural speech interactions, task requests are identified via  keyword-matching while an LLM (GPT-4o) handles intent recognition for social speech.
In this work, we used the Panda robot from Franka Emika, an industrial robotic arm with 7 DoF. 

\subsection{Speech Listener: Continuous Speech Processing}
The listener captures audio from the participant's microphone, 
converting speech into text to identify both task requests and opportunities for small talk. 
It runs in the background throughout the interaction. 
Using the Google Cloud Speech-to-Text (STT) API, the listener processes audio in 2.6-second chunks at a 16 kHz sampling rate.
If silence is detected throughout a chunk, the API determines that a final transcript has been obtained. 
The listener then pauses momentarily to prevent confusion between the robot's and user’s speech, and resumes listening after the robot delivers its speech response. 

A monitor between the robot and user displays the system's current status---whether it is actively listening, processing speech, or pausing during the robot’s response. 
A ding sound effect signals each transition between listening and intent recognition. 
After processing the STT, the system enters two phases: intent recognition and robot behavior generation.

\subsection{Phase 1: Intent Recognition}
To address the variable response times of the OpenAI API calls in real-time HRI, we separate task request identification from social speech processing. 
Specifically, the partial and final transcript results are fed into our system's request recognizer to expedite the process of recognizing a task request.


\textbf{Recognizing Task Requests.} 
For task requests, we perform local detection using a sliding window approach, scanning for key words like COLOR (yellow/green) and PIPE in the user’s speech. 
By running the request recognizer on both the partially generated and final transcripts, we aim to improve responsiveness. 
For instance, if the OpenAI API response generation is delayed, we can still identify the task request first, allowing the system to proceed with task-oriented responses and actions, thereby reducing perceived wait times.
Thus, our system always speaks task-oriented language before engaging in any small talk responses.

We employed the GPT-4o model to handle both social intent recognition and response generation. 
To efficiently interpret a user's social speech and generate a conversational response, the final transcript is processed through the OpenAI API, which returns a JSON response based on our custom prompt (See Appendix B). 
This prompt incorporates a user's general interest question obtained from the pre-study survey to tailor the robot's small talk to their interests. 
The model has access to the entire conversation history, including the user's STT inputs and the robot’s TTS outputs, ensuring fluid conversations and enabling the model to decide when to shift conversation topics. 
The model identifies social speech by determining which parts of the user's input are not task-oriented and generates a small talk reply based on this content.
The JSON output includes whether small talk was detected, the excerpt of the user's social speech, and the generated response.
While every call to the model in \textit{Phase 1} generates a small talk reply, whether or not it is used depends on the decisions made in \textit{Phase 2}, which governs when the system should initiate small talk.

\subsection{Phase 2: Robot Behavior Generation}
\textbf{Task Behaviors.} 
The system randomly selects a confirmation message from a bank of pre-written phrases if a pipe request is detected in the user's speech and executes the message using the Google Cloud Text-to-Speech API. 
We chose to use pre-written phrases to streamline task-oriented processes and reduce the perception of delays. 
Simultaneously, the system triggers a pre-programmed ROS script to retrieve the corresponding pipe from a pipe dispenser. 
These ROS actions run as background processes, allowing the user to continue communicating with the robot while it moves. 
To ensure the robot only handles one pipe retrieval at a time, if a participant requests a pipe while the robot is still moving, the robot responds with, ``I must finish my current action before processing a new request.''
However, the robot can still engage in small talk if applicable. 

\textbf{Small Talk Behavior.}
For small talk generation, we consider both the presence of a pipe request and the user's social speech to adjust the robot's small talk frequency.
The robot initiates small talk only after every other pipe request but can respond whenever it detects social speech in the user's input. 
To help users get accustomed to the main task, the system is programmed not to initiate small talk until after the third pipe request. 
However, if a user initiates small talk before this point, the robot is allowed to respond. 
Once small talk is triggered, the reply generated by the language model is spoken via text-to-speech (TTS).
Although our system logic controls when to start small talk, the system always waits for the user to conclude the conversation.



\section{Methods}
We conducted an experiment to examine the effects of small talk on task efficiency, rapport, and interaction dynamics. 
\subsection{Experimental Conditions and Task}
 
Participants were randomly placed in one of two conditions:
\begin{itemize}[leftmargin=*]
    \item \textbf{Functional (F).} The robot engaged exclusively in task-oriented speech by confirming participant requests.
    \item \textbf{Social (S).} In addition to task speech as in the functional condition, this robot initiated small talk with participants and responded to their small talk remarks following the implementation described in Section \ref{sec:robot_system}.
\end{itemize}
Participants collaborated with either robot to complete a PVC pipe assembly task.
The participant and robot had their own workspace and access to different assembly pieces; the participant needed to verbally request yellow and green pipes from the robot to complete the task.
\subsection{Hypotheses}
\textbf{H1.} Participants in the social condition will spend more time completing the task than those in the functional condition.

\textbf{H2.} Participants in the social condition will experience higher levels of rapport than those in the functional condition.

\textbf{H3.} Participants in the social condition will experience less boredom than those in the functional condition. 

\textbf{H4.} Participants in the social condition will want to work again with the robot more than in the functional condition.

\subsection{Procedure}
Participants first completed a consent form and two pre-study surveys.
The first survey gathered demographic information, assessed experience with voice assistants and robots, and included a general interest question (\eg Travel, Music, Sports). 
Responses to this question were incorporated into the system prompt to slightly tailor the robot's small talk.
The second survey was an abridged Five Factor Model questionnaire, collecting only extroversion data on a 1--7 scale. 
The experimenter provided a brief task overview and written instructions. 
Participants completed a practice task to familiarize themselves with the robot’s task-oriented behaviors and use of verbal commands. 
In the main task, participants assembled a more complex PVC structure.
The task began when they pressed a green button, and a ping sound indicated when the robot began processing the speech input.
No additional speech input was accepted during processing.
After completing the structure, participants pressed a red button to end the task.
A final survey and semi-structured interview followed.
The study, approved by our IRB, lasted about 45 minutes. Participants were compensated at a rate of \$15 US dollars per hour.

\section{Measures}

We used four primary metrics to evaluate our main hypotheses. 
Additionally, we included metrics for exploratory analyses without specific hypotheses in order to provide further insights into participants' behaviors and affective expressions during the interaction.
All time metrics are reported in minutes.
To assess our experimental manipulation, we used two yes-no questions in the post-study questionnaire: ``The robot spoke to me'' and ``The robot spoke to me only about the pipe task.'' 

\subsection{Primary Metrics}

\textbf{1. Task Duration.}
The time a participant took to complete the task, from when they pushed the green button to indicate start to when they pressed the red button to indicate end.

\textit{All constructs below are on a 1--7 scale. Appendix A
includes questionnaire items for each scale construct.}

\textbf{2. Rapport }\textit{(Four items; Cronbach's $\alpha = 0.83$).}
This scale sought to measure the perceived rapport between the user and robot built during the task interaction.

\textbf{3. Boredom} \textit{(One item).}
We assessed participants’ boredom with the question, ``I felt bored while completing the task.''

\textbf{4. Longer Working Relationship} \textit{(Two items; Cronbach's $\alpha = 0.86$).}
This scale sought to assess a user's willingness to work with the robot over a longer period of time.

\subsection{Exploratory Metrics}

\textit{1) Task Time Breakdown.} 
We investigate criteria that may have impacted overall task duration.

 \begin{itemize}[leftmargin=*]
      \item \textbf{(Human/Robot) Active Working Time.} 
      The total time a human or robot spent actively working on the task. 
      The human's active time was computed by summing the annotated human active working time.
      The robot's active time was computed by summing the times from when it began executing a pipe request to when it finished.

      \item \textbf{(Human/Robot) Idle Time.} 
      This is the total time a human or robot spent \textit{not} actively working on the task. 
      It was calculated by subtracting a human or robot's active working time from their overall task duration. 

\end{itemize}

\textit{2) Speech Duration.}
We examine the participants' task-oriented and social speech duration during the task.

\begin{itemize}[leftmargin=*]

    \item \textbf{Total (Task/Social) Speech Time}. 
    We identify the number of words in participants' speech and estimate the duration of their total speaking time using the average speaking speed rate of 150 words per minute (wpm).
    \item \textbf{Averaged (Task/Social) Speech}.
    This was calculated as total task-oriented or social speaking time over \textit{task duration}.


\end{itemize}

\textit{3) Verbal Response Rates.}
How frequently participants engaged with the robot's verbal communication. All are calculated from the transcripts' counted labels (See Appendix C).

\begin{itemize}[leftmargin=*]
 
    \item \textbf{(Task/Social) Response Rates} ($\%$). 
    We calculate the percentage of the robot's speech phrases that received a verbal response from the user. 
    We categorize a robot's phrases into task and social speech (See Appendix C). 

\end{itemize}

\textit{4) Small Talk Interaction Dynamics.}
\label{results:st-dynamics}
How participants socially engaged with the robot's verbal communication; conversation characteristics related to initiation and turn-taking.

\begin{itemize}[leftmargin=*]
    \item \textbf{First Response to Robot Small Talk.} 
    We observe when the participant made their first response to the robot's small talk speech.
    We counted the number of robot small talk attempts made until the user responded to the robot.

    \item \textbf{Number of Participant-Initiated Questions.} 
    We count the number of questions initiated to the robot by a participant.

    \item \textbf{Longest Small Talk Exchange.} 
    Additionally, we identify the most prolonged small talk exchange between the robot and the participant. 
    This is the longest number of turns between two pipe requests and does not account for any exchanges after the last pipe has been retrieved.

    \item \textbf{Lingering Conversations.}
    We observed if a continued small talk exchange existed at the end of the task; we counted the number of human speech turns after the last pipe request.

    \item \textbf{Farewell Exchanges.}
    We observed if a participant exchanged a farewell or ``goodbye" with the robot; we counted the number of participants that engaged in this behavior.

\end{itemize}

\begin{figure*}[ht!]
    \centering
    \includegraphics[width=.9\textwidth]{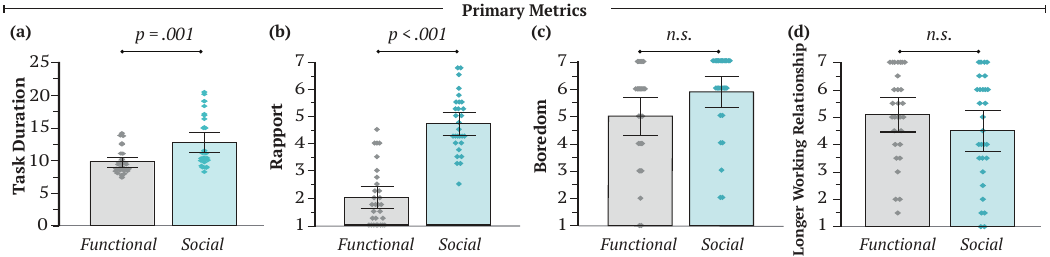} 
    \caption{We investigated four primary metrics;
    a significant difference was found between conditions for (a) task duration and (b) perceived rapport.}
    \label{fig:primary} 
\end{figure*}

\textit{5) Affective Expressions.}
Finally, we measured the participants' affective responses (See Section \ref{sec:analysis}).
\begin{itemize}[leftmargin=*]

    \item \textbf{Emotional Expressiveness.} 
    We report the average percentage of each emotion present during the interaction, dividing 
    the number of frames where a specific emotion was detected
    by
    the number of frames in which a face was identified.

    \item \textbf{Overall Emotional Expressiveness.} 
    We observe the total overall expressiveness of a participant during the task interaction. 
    This is computed by combining all the individual emotional expressiveness of a participant. 
    
\end{itemize}

\subsection{Data Analysis}
\label{sec:analysis}
We gathered system logs and audio-visual recordings of participants interacting with the robot during the experiment. 
These data were processed and used in our data analysis.


\textit{Human Active Working Time.}
We labeled the time a participant was actively working on the pipe task. 
After familiarizing ourselves with the data, we created criteria to help distinguish whether a participant was actively working on the task. A primary coder labeled data for all participants, and a secondary coder annotated 10\% of the data. After resolving disagreements, we found almost perfect inter-coder agreement on the data; Intraclass Correlation Coefficient (ICC3) of 0.99.

\textit{Affective Expressions.}
We analyzed the video data from a participant-facing webcam via OpenFace2.0 \cite{openFace2}. 
Facial action units (AUs) \cite{openFaceAU} were first extracted and used to detect if a specific emotion is present (See Appendix D). 
We detected the presence of Happiness/Joy, Disgust, Sadness, Anger, Fear, Contempt, and Surprise  \cite{friesen1983emfacs}. 
We only selected video frames that detected a face. 
For frames with multiple emotions detected, the highest-intensity emotion was selected as the predominant emotion. 
If there was only one AU or less detected from the OpenFace system output and major emotion was not identified, we labeled this frame as ``Neutral.'' 
Emotional presence was normalized across the task duration.

\subsection{Participants}
Seventy participants were recruited through a community mailing list and local flyers. 
Twelve were excluded due to various issues; 
one was excluded due to having prior knowledge of the experiment details; 
two were excluded for not following task instructions; 
five were excluded due to the system's speech-to-text software having difficulty understanding their native accent; 
four were excluded due to technical issues (\eg malfunctions with speakers, robot gripper, and corrupted logging script).
Of the remaining participants ($N=58$), each condition had 29 participants (15 male, 14 female). 
Participants had an average age of 25.12 ($SD=6.55$). 
On a scale from 1 to 7, their mean scores for 
personality (extroversion), 
experience with voice assistants,
and 
experience with robots were
3.88 ($SD=1.43$), 
4.86 ($SD=1.66$),
and 
3.33 ($SD=1.67$), 
respectively; there were no significant differences between conditions. See Appendix G for additional participant demographics details.

\section{Results}
All participants passed our manipulation checks. 
Fig.
\ref{fig:primary}
-
\ref{fig:task-time-expressions} 
summarize key results. 
We tested for normality and used standard t-tests; 
if data was non-normal, we used Welch's Test or the Wilcoxon Signed-Rank Test for matched pairs.

\subsection{Primary Metrics}

\textbf{1. Task Duration.}
Participants in condition S spent significantly more time finishing the task than those in condition F, $t(40.48) = 3.55, p = .001$

\textbf{3. Rapport.} Participants in condition S reported significantly higher levels of rapport with the robot compared to those in condition F, $t(56)=-9.48, p < .001$.

\textbf{2. Boredom.} There were no significant differences in task boredom reported by participants across the two conditions, $t(56) = -1.98, p = .05$.

\textbf{4. Longer Working Relationship.} There were no significant differences between conditions, $t(56)=1.24, p = .22$ (For primary metrics $1$-$4$, see Fig. \ref{fig:primary}). 

\subsection{Exploratory Metrics}

\begin{figure*}[ht!]
    \centering
    \includegraphics[width=\textwidth]{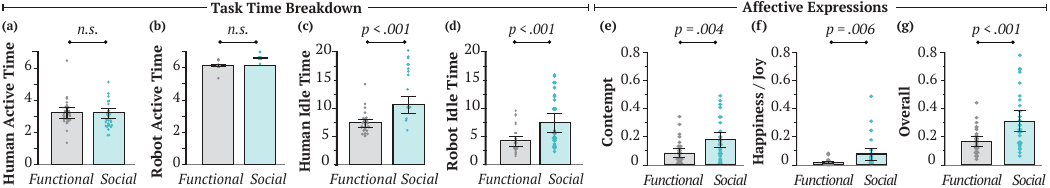} 
    \caption{We plot task time breakdown results by condition in Fig. \ref{fig:task-time-expressions}a-\ref{fig:task-time-expressions}d. Significant differences were found for (c) human idle time and (d) robot idle time.
    We also plot the significant differences between conditions for individual and overall emotional expressiveness in Fig. \ref{fig:task-time-expressions}e-\ref{fig:task-time-expressions}g.
    }
    \label{fig:task-time-expressions} 
\end{figure*}

\textit{1) Task Time Breakdown}
\begin{itemize}[leftmargin=*]
    \item \textbf{(Human/Robot) Active Working Time.}
    We found no significant difference between participants' active time across conditions, 
    $t(56) = 0.01, p =.990$.
    Similarly, we found no significant difference between the robot's active time across conditions, 
    $t(56) = 0.30, p =.762$ (See Fig. \ref{fig:task-time-expressions}a and \ref{fig:task-time-expressions}b).

    \item \textbf{(Human/Robot) Idle Time.}
    Participants had significantly higher idle time in condition S than in condition F, $t(40.26) = 3.86, p < .001$.
    Likewise, the robot spent more time in the idle mode in condition S than in condition F, $t(41.04) = 3.56, p < .001$ (See Fig. \ref{fig:task-time-expressions}c and \ref{fig:task-time-expressions}d).
\end{itemize}


\textit{2) Speech Duration}

\begin{itemize}[leftmargin=*]
    \item \textbf{Total (Task/Social) Speech Time.}
    There was a significant difference for participants' total task speaking time between condition F ($M=0.54, SD = 0.05$) 
    and S ($M=0.67, SD = 0.12$), $t(38.15) = 5.25, p <.001$.
    We found the total social speaking time across participants in condition S as 1.59 min
    ($SD = 1.24$). 
    No participants engaged in social speech with the robot in condition F.

    \item \textbf{Averages (Task/Social) Speech.}
    For participants in the social condition,
    a Wilcoxon Signed-Rank Test found a significant difference between the means of (social speech / task duration) and (task speech / task duration), 
    suggesting that participants engaged in significantly more social than task speech  ($p < .001$).

\end{itemize}

\textit{3) Verbal Response Rates}
\begin{itemize}[leftmargin=*]

    \item \textbf{(Task/Social) Response Rates.}
    Table \ref{tab:response-rates} summarizes participants' response rates to a robot's task and social speech.
    All social condition participants engaged in small talk. 
    
\end{itemize}

\begin{table}[ht!]
\centering
\caption{Task and Social Response Rates by Condition}
\label{tab:response-rates}
\resizebox{\columnwidth}{!}{%
\begin{tabular}{rcrcrc}
\multicolumn{1}{c}{\textbf{User Response Type}} & \textbf{Cond.} & \multicolumn{2}{c}{\textbf{Count}} & \multicolumn{2}{c}{\textbf{People}} \\ \hline \hline

\multirow{2}{*}{To Robot Confirmation} & F & 8 / 471 & 2\% & 4 / 29 & 14\% \\ \cline{2-6} 
 & S & 15 / 473 & 3\% & 9 / 29 & 31\% \\ \hline
To Robot Question & S & 729 / 838 & 87\% & 29 / 29 & 100\% \\ \hline
To Robot Statement & S & 42  / 68 & 62\% & 14 / 29 & 48\%
\end{tabular}%
}
\end{table}

\textit{4) Small Talk Interaction Dynamics}

\begin{itemize}[leftmargin=*]
    \item \textbf{First Response to Robot Small Talk.}
    72\% of our participants ($n=21$, out of 29) responded to the robot's first small talk question.
    21\% ($n=6$) did not engage in small talk until the robot's second question. 
    The remaining two participants waited longer to engage, not responding to the robot until its sixth and tenth small talk attempts, respectively.

   \item \textbf{Number of Participant-Initiated Questions.}
   59\% of our participants ($n=17$, out of 29) initiated 99 questions to the robot. The number of questions initiated per person ranged between 0 and 29 ($M=3.41,Mdn=1,SD=6.02$).
   Only seven questions, asked by five unique participants, were intended to clarify or ask the robot to repeat what it had said. 
   Thus, 52\% of our participants ($n=15$, out of 29) 
   asked 92 non-clarification questions. 

   \item \textbf{Longest Small Talk Exchange.}
   The longest small talk exchange in number of turns per participant ranged between 2 and 14 ($M=4.47,Mdn=3,SD=2.87$).

   \item \textbf{Lingering Conversations.}
   73\% of our participants ($n=19$, out of 29) engaged in small talk with the robot after the final pipe request.
   The number of turns taken ranged between 0 to 23 ($M=3.69, Mdn=2, SD=5.41$).  
   Fig. \ref{fig:teaser} illustrates an example of such lingering conversation.

    \item \textbf{Farewell Exchanges.}
    59\% of our participants 
    ($n=17$, out of 29) engaged in a farewell exchange with the robot at the end of their interaction. 
    This included eight participants who had \textit{lingering conversations} with the robot.

\end{itemize}

\textit{5) Affective Expressions}

\begin{itemize}[leftmargin=*]
    \item \textbf{Emotional Expressiveness.} 
    We detected higher levels of emotional expressiveness 
    in
    condition S than 
    condition F
    for
    contempt, 
    happiness/joy, 
    fear, and 
    surprise. 
    We found significant differences between conditions for
    contempt ($t(45.79) = 3.0, p = .004$), 
    happiness/joy ($t(29.72) = 2.95, p = .006$),
    and 
    fear\footnote{Exact fear mean values are close to zero. Condition S: $M=0.00024, \\SD=0.00031$. Condition F: $M=0.0001, SD=0.0001$.} ($t(38.46) = 2.21, p = .034$). 
    Differences between conditions for
    disgust, 
    anger,  
    surprise,
    sadness,
    and neutral were not significant (See Appendix E).

    \item \textbf{Overall Emotional Expressiveness.} 
     When comparing the means of the total emotions detected 
     in condition F ($M=0.16, SD=0.09$) 
     and
     condition S ($M=0.31, SD=0.21$), 
     Welch's Test indicated a significant difference between conditions $t(39.29) = 3.58, p < .001$; 
     participants in condition S were more expressive than those in condition F (Fig. 
     \ref{fig:task-time-expressions}g).

\end{itemize}

\section{Discussion}
This work presents a robotic system capable of engaging in small talk with people and examines the effects of small talk from a non-anthropomorphic manipulator on a physically collaborative task. 
Below, we discuss the benefits and drawbacks of integrating small talk in collaborative robots.


\subsection{Effects of Novelty on Task Efficiency}
Participants in the social condition spent longer time on task compared to those in the functional condition, \textbf{\textit{confirming H1}}.
However, it is important to note that the increased task time was due to ``idle'' rather than active working time, which was similar across conditions.
Thus, we infer that small talk did not directly impact active working time on the task.
S47 explained, 
\textit{``I was still working while just talking. Those two things don't really distract from one another so I thought it was fine''}. 

The extended idle time measured could potentially be attributed to the novelty of the interaction. 
Participants in the social condition were at first surprised, as they were unaware that the robot had social capabilities.
S12 noted, 
\textit{``I just was really thrown off when it started asking me stuff...''}. 
R41 shared, 
\textit{``I was ... super surprised that it was talking to me initially ... after I got the hang of responding to it and asking it to bring stuff at the same time, it went ... just as smooth''}. 
This suggests that novelty effects \cite{reimann2023social} may have contributed to increased idle time spent chatting with the robot.
Our findings highlight the need for future work to explore how these effects may persist or change over time. 


\subsection{Small Talk Improves Rapport}
Participants reported significantly higher rapport ratings in the social condition than the functional condition, \textbf{\textit{confirming H2}}. 
The greater perceived rapport may be attributed to social interactions, as all 29 participants in the social condition engaged in small talk with the robot, and 93\% responded within the robot's first or second small talk attempt.
Moreover, 59\% of participants in the social condition initiated questions to the robot, with the majority being meaningful and extending beyond basic clarification questions (See Section \ref{results:st-dynamics}).
This inclination to both respond and ask questions suggests a genuine interest in and substantial engagement with the robot. 
In fact, for some participants, their \textit{longest small talk exchange} and \textit{lingering conversation} lasted 14 and 23 turns, respectively (See Section \ref{results:st-dynamics}).
Another participant, who had their \textit{longest small talk exchange} lasting six turns and a \textit{lingering conversation} of 15 turns, explained that their curiosity drove them to continue interacting with the robot: 
\textit{``I wanted to see the robot's response to questions, so I kept talking to it''} (S47), offering additional evidence of the novelty effect.

Furthermore, of the participants who engaged in a farewell exchange with the robot, 
some exchanges were brief, such as S31 saying, \textit{``Thank you''}, to which the robot responded,\textit{``You're welcome. Have a great day.''}
Others were more personal and supportive; S45 said, \textit{``Thank you. You're doing a great job. You can rest now''}, followed by an exchange of \textit{``Goodbye''} between the robot and participant. 
Another participant, who appeared reluctant to end the task immediately, even used the expression, ``See you later, alligator'' (Fig. \ref{fig:teaser}). 
These farewell exchanges, along with long stretches and lingering conversations, support the development of rapport between people and the robot. 
Future research should study the possible benefits of rapport, such as tolerance of robot errors, in repeated human-robot collaboration.



\subsection{Monotonous Task Trumps Small Talk}
We did not find statistical evidence to support H3 and H4.
Nevertheless,
small talk seems capable of making tedious tasks more enjoyable.
S13 explained, \textit{``It was fun. It's a very monotonous task ... simple task, but I think I won't be able to do this for hours''}. 
Similarly, S41 shared how the small talk \textit{``made it more fun, I guess, to put stuff together''}.
We recommend exploring how to design social features in collaborative robots to help alleviate boredom and enhance user engagement in repetitive tasks, common in workplaces where collaborative robots are deployed. 
This investigation could address 
employee job dissatisfaction and reduced motivation during long, tedious shifts in warehouses or industrial settings \cite{ingsih2021role}. 
These conditions often lead to diminished performance and high turnover rates for companies \cite{autry2003warehouse}. 
Increasing employee job satisfaction could reduce turnover rates and boost productivity by retaining skilled employees and lessening the need to train new workers repeatedly.

\subsection{Reflecting on the Design of Small Talk}

In this work, we used the generative power of a large language model to drive the robot's small talk; we additionally tailored the model's behavior to participants by incorporating their general interests in our prompt for conversation topics. 
The combination of the generative capabilities of an LLM and our tailored topic selection
led the participants to characterize the robot as 
\textit{``very cool''} (S12)  and \textit{``personable''} (S24).
R64 noted, 
\textit{``It kept layering the questions, going deeper. That made the experience... more human-like''}.
Participants also described their interaction with the robot as  \textit{``enjoyable``} (S3), \textit{``pretty fun''} (S9), and \textit{``quite pleasant''} (S23).


Despite the overall positive experiences, 
participants indicated the need to fine-tune the robot's small talk behavior, especially regarding its timing and frequency.  
In this work, we limited the robot’s initiation of small talk to every other pipe request while allowing it to respond if any small talk in the user's speech is detected.
This design led some participants to perceive the robot's responses as excessive. 
S23 commented, 
\textit{``I like the kinds of questions ... but the intervals that it asked me was a bit too soon''}, 
and described the robots' follow-up questions as 
\textit{``constant''}.
S68 expressed a similar sentiment, stating, 
\textit{``it would just keep shooting questions''}. 

Future work should aim to strike the right balance with a robot's small talk timing and frequency.
One approach could use real-time 
emotion analysis \cite{Lu_Huang_Lee_2024}.
Our study found significantly more emotional presence in the social condition than in the functional condition.
Future work could explore using real-time facial expressions to dynamically adjust the timing and frequency of small talk.
Additionally, identifying user states (\eg active or idle) in real time could further enhance small talk adaptation to balance rapport and productivity effectively.

\subsection{Do We Really Want Human-Robot Small Talk at Work?}

Considering our findings, 
we believe that integrating small talk into human-robot collaboration is still worthwhile.
People are already accustomed to small talk among human co-workers. 
One participant shared, 
\textit{``I had worked in big companies... and I really enjoyed the people around me... that bantered. The jokes, the water cooler talk...''}(S23). 
Even in industrial manufacturing environments, 
operators engage in casual conversations or ``small talk'' with colleagues during shifts \cite{cheon2022robots}, and some have found themselves ``talking'' to the robots they work with.
Small talk could be incorporated at the beginning or end of manufacturing shifts or for adjacent human-robot workstations, similar to two side-by-side human workers engaging in ST \cite{Sauppé_Mutlu_2015}.
One participant explained their motivation for prolonged interaction with the robot, saying,
\textit{``I had... the thought of this is the last time I'll talk to a robot again''}(S58). 
This suggests an inherent human desire for social interaction, even with non-anthropomorphic robots. 
Another participant noted, 
\textit{``If I was like a cog in the machine and I had to do this day in and day out, like make these things, I feel like eventually I would want to talk to something, but I would prefer to talk to a person... if that wasn't an option, then I would want to talk to the robot''} (S51).
While we are not proposing to replace human-human small talk, 
we believe that if designed correctly,
human-robot small talk can be acceptable, beneficial, and desirable in collaborative settings.

\subsection{Ethical Implications}
From our experiences integrating LLMs into robots designed to interact with people, we have learned important lessons and urge researchers to consider the ethical implications of these systems carefully. 
One participant perceived the robot's response as empathetic after sharing that they misplaced vacation photos, stating 
\textit{``The empathy that it showed was quite nice... I think I said the word, `sorry'. It said `say it's OK. Memories are all that count'. So I thought that's really intelligent of the robot to ... make that connection ...''} (S23). 
Such responses raise concerns about potential deception in human-subjects research.
While the CASA paradigm \cite{nass1994computers} illustrates that people respond to machines in inherently social ways, LLMs present a unique challenge as their sophisticated behavior can lead people to attribute even greater human characteristics to them compared to traditional machines.

Furthermore, some participants felt a sense of ``obligation'' to respond to the robot. 
One participant, who engaged in nine turns for their \textit{longest small talk exchange} said:
\textit{``Towards the end, I wanted to get it done... I don't know why I didn't ... I could have just been like, no, give me the thing''} (S68, Appendix F).
S41, who had a \textit{lingering conversation} of ten turns, expressed feelings of guilt:
\textit{``It just kept asking questions at the end... I feel bad stopping it since the task was done''}. 
These accounts illustrate how an LLM-powered system, especially with an embodiment, could create unwanted social pressure and influence people's behavior.


Additionally, when the system was prompted to ``generate other information about [itself] based on these [fun] facts'' (See Appendix B), 
it sometimes fabricated details that could mislead participants. For example, when asked, 
\textit{``Are you the robot that I see when I walk by the archway like a building down?''} (S38), 
the robot replied, 
\textit{``Yes, that's likely me. I'm often around here.''} 
despite not knowing the location. 

To address these concerns, we recommend the following:
\begin{itemize} [leftmargin=*]
    \item \textbf{Obtain IRB approval for potential deception in LLM-integrated human-robot interaction studies.} Researchers should inform participants about possible deception in advance or, at very least, ensure thorough debriefing afterward.
    \item \textbf{Address potential misinformation in post-study debriefing.} During debriefing, researchers should clarify that the robot’s responses may not be accurate and advise participants to verify any information or advice received. 
\end{itemize}

\section{Limitations}
This study used a simulated task in a short-term interaction primarily with university students and employees. 
Future research should examine small talk with operators in real industrial environments and study longitudinal small talk interactions.
This could also uncover potential effects of time pressure and task severity alongside small talk.
While we did not conduct a cultural analysis, we urge researchers to explore cross-cultural and language-related aspects using empirical data.



\section{Conclusion}
Our study provides initial evidence of the benefits and challenges of incorporating small talk into non-anthropomorphic robots commonly used in industrial settings.
While participants engaged with and generally appreciated the robot’s small talk attempts, attributing the robot traits like empathy and intelligence, these interactions raise ethical concerns, including the potential for deception and unintended social pressure to engage. 
Although we are not suggesting that human-robot small talk should replace human-human interactions, our results support the belief that, if crafted carefully, it can be valuable and welcomed in collaborative settings. 
Strategies to balance the frequency and timing of robot-initiated small talk could further enhance user experiences. 
This work is a step toward harnessing the potential of small talk in human-robot collaboration. 



\section*{Acknowledgment}
This work was supported by National Science Foundation award \#2141335. 

\newpage




\bibliographystyle{IEEEtran}
\balance
\bibliography{IEEEabrv,references}


\end{document}